\documentclass{article}

\usepackage{amsmath}
\usepackage{amsfonts}
\usepackage{amssymb}
\usepackage{color}  
\usepackage{amsthm}
\usepackage{graphicx}
\usepackage{caption}
\usepackage{comment}
\usepackage{subcaption}
\usepackage{algpseudocode}
\usepackage{algorithm}
\usepackage[utf8]{inputenc}
\usepackage{tabularx}
\usepackage{multirow}
\usepackage{hyperref}

\DeclareMathOperator*{\argmin}{arg\,min}

\usepackage{natbib}
\bibpunct[, ]{(}{)}{,}{a}{}{,}%

\title{Integer Programming for Multi-Robot Planning: \\ A Column Generation Approach}

\author{Naveed Haghani\textsuperscript{\rm 1}, Jiaoyang Li\textsuperscript{\rm 2}, Sven Koenig\textsuperscript{\rm 2},\\ Gautam Kunapuli\textsuperscript{\rm 3}, Claudio Contardo\textsuperscript{\rm 4},  Julian Yarkony\textsuperscript{\rm 3}\\[2ex] 
\textsuperscript{\rm 1}University of Maryland, College Park, MD\\ 
\textsuperscript{\rm 2}University of Southern California, Los Angeles, CA\\ 
\textsuperscript{\rm 3}Verisk Computational and Human Intelligence Laboratory, Jersey\\
\textsuperscript{\rm 4}ESG UQAM and GERAD, Montreal, Canada
 City, NJ\\ 
}\date{June 2020}

\begin{document}

\maketitle
\begin{abstract}

We consider the problem of coordinating a fleet of robots in a warehouse so as to maximize the reward achieved within a time limit while respecting problem and robot specific constraints.  We formulate the problem as a weighted set packing problem where elements are defined as being the space-time positions a robot can occupy and the items that can be picked up and delivered.  We enforce that robots do not collide, that each item is delivered at most once, and that the number of robots active at any time does not exceed the total number available.  Since the set of robot routes is not enumerable, we attack optimization using column generation where pricing is a resource-constrained shortest-path problem.

\end{abstract}

\section{Introduction}
In this paper, we tackle multi-robot planning (MRP), which aims to route a fleet of robots in a warehouse so as to achieve the maximum reward in a limited amount of time, while not having the robots collide and obeying the constraints of individual robots. In MRP, individual robots may make multiple trips over a given time window and may carry multiple items on each trip.  We optimize the efficiency of the warehouse, not the makespan, since we expect new orders to be continuously added.  Our contributions are that (1) we  adapt the integer linear programming (ILP) formulation and column generation (CG) approach for (prize collecting) vehicle routing \citep{Desrochers1992,stenger2013prize} to MRP and (2) adapt the seminal work of \citep{boland2017continuous} to permit efficient optimization by avoiding consideration of every time increment.  

Routing problems for a fleet of robots in a warehouse are often treated as Multi-Agent Pathfinding problems (MAPF) \citep{SternSoCS19}.  In MAPF, we are provided with a set of agents, each with an initial position and destination.  The goal is to minimize the sum of the travel times from the initial position to the destination over all agents such that no collisions occur. MAPF can be formulated as a minimum cost multi-commodity flow problem on a space-time graph \citep{yu2013planning}.  Optimization can be tackled using multiple heuristic and exact approaches, including search \citep{LiICAPS2020}, linear programming \citep{yu2013planning}, branch-cut-and-price \citep{lam2019branch}, satisfiability modulo theories \citep{SurynekIJCAI19}, and constraint programming \citep{GangeICAPS2019}.

One common shortcoming in MAPF approaches is that they require that robot assignments be set before a robot route can be solved.  The delegation of robot assignments and the optimal set of routes for the fleet are treated as independent problems.  Several recent works~\citep{MaAAMAS17,LiuAAMAS19,GrenouilleauICAPS19,FarinelliAAMAS20} solve this combined problem in a hierarchical framework, i.e., assigning tasks first by ignoring the non-colliding requirement and then planning collision-free paths based on the assigned tasks.  However, these methods are non-optimal as the consideration of possible collisions can easily affect the optimal task assignment for the fleet.

We organize this paper as follows. In Section \ref{setPackForm}, we formulate the MRP as an ILP, which we attack using CG in Section \ref{CGSec}.  
In Section \ref{Pricing}, we solve the corresponding pricing problem as a resource-constrained shortest-path problem (RCSP).  In Section \ref{doi_sec}, we provide dual optimal inequalities, which accelerate CG.  In Section \ref{exper}, we demonstrate the effectiveness of our approach empirically.  In Section \ref{conc}, we conclude and discuss extensions.

\section{Problem Formulation}
\label{setPackForm}
In this section we present the problem we call Multi-Robot Planning, outlining it and then formulating it as an ILP.  We conclude with detailing feasibility constraints and cost associations for robot routes.  We are given a fleet of mobile warehouse robots that enter the warehouse floor from a single location, called the launcher, pick up one or multiple items inside the warehouse, and deliver them to the launcher before the time limit. Each item has a reward (i.e., negative cost) and a time window during which the item can be picked up. Each robot has a capacity and is allowed to perform multiple trips. At the initial time, the fleet of robots is located at the launcher, however we also allow for some robots, called extant robots, to begin at other locations. The use of extant robots permits re-optimization as the environment changes, e.g. when items have their rewards changed or when items are added or removed. Our goal is to plan collision-free paths for the robots to pick up and deliver items and minimize the overall cost.

For computational efficiency, we approximate the continuous space-time positions that robots occupy by treating the warehouse as a 4-neighbor grid and treating time as a set of discrete time points. Each position on the grid is referred to as a cell.  Most cells are traversable for the robot but some cells are labeled as obstacles and cannot be traversed, we call these obstructed.  Through each time period, robots are capable remaining stationary or moving to an adjacent unobstructed cell in the four main compass directions, which we connect through edges.  Robots are required to avoid collisions by not occupying the same cell at any time point and not traversing the same edge in opposite directions between any successive time points.  Every item is located at a unique cell.  Robots incur a cost while deployed on the grid and for moving on the grid, however they can obtain a reward for servicing an item.  To service an item, a robot must travel to the specific cell where the item is located during the item's associated serviceable time window.  Servicing an item consumes a portion of the robots capacity, which can be refreshed once it travels back to the launcher.


%
We formulate MRP as an ILP problem using the following notation.  We use $\mathcal{G}$ to denote the set of feasible robot routes, which we index by $g$. We note that $\mathcal{G}$ is too large to be enumerated.  We use $\Gamma_g \in \mathbb{R}$ to denote the cost of robot route $g$.  We  use $\gamma_g \in \{0,1\}$ to describe a solution where $g$ is in the solution IFF $\gamma_g=1$. We describe the sets of items, times, and extant robots as $\mathcal{D}$, $\mathcal{T}$, and $\mathcal{R}$, respectively, which we index by $d$, $t$, and $r$, respectively.  
We use $(\mathcal{P},\mathcal{E})$ to denote the time-extended graph. Every $p \in \mathcal{P}$ represents a space-time position, which is a pair of a location (i.e., an unobstructed cell on the warehouse grid) and a time $t \in \mathcal{T}$. Two space-time positions $p_i, p_j \in \mathcal{P}$ are connected by a (directed) space-time edge $e=(p_i, p_j) \in \mathcal{E}$ IFF the locations of $p_i$ and $p_j$ are the same cell or adjacent cells and the time of $p_j$ is the time of $p_i$ plus one.

We describe routes using $G_{ig} \in \{0,1\}$ for $i \in \mathcal{I}= \{ \mathcal{D} \cup \mathcal{T}\cup \mathcal{P}\cup \mathcal{E}\cup \mathcal{R} \}$.  We set $G_{dg}=1$ IFF route $g$ services item $d$.  We set $G_{tg}=1$ IFF route $g$ is active (meaning moving or waiting) at time $t$.  We set $G_{pg}=1$ IFF route $g$ includes space-time position $p$. We set $G_{rg}=1$ IFF route $g$ is associated with extant robot $r$.  We set $G_{eg}=1$ IFF route $g$ uses space-time edge $e$. This edge is associated with adjacent cells $e_1$ and $e_2$ in space and time $t$. Thus, $G_{eg}=1$ indicates that a robot on route $g$ crosses from $e_1$ at time $t$ to $e_2$ at time $t+1$ OR from $e_2$ at time $t$ to $e_1$ at time $t+1$. We use $N$ to denote the total number of robots available in the fleet.  We write MRP as an ILP as follows, followed by an explanation of the objective and constraints.

\begin{algorithm}[!b]
 \caption{Optimization via Column Generation}
\begin{algorithmic}[1] 
\Repeat
\State $\gamma,\lambda \leftarrow $ Solve the RMP  over $\hat{\mathcal{G}}$
\State $g^* \leftarrow \min_{g \in \mathcal{G}}\bar{\Gamma_g}$
\State $\hat{\mathcal{G}} \leftarrow \hat{\mathcal{G}} \cup \{ g^*\} $
 \Until{ $\bar{\Gamma}_{g^*} \geq 0$ }
 \State $\gamma \leftarrow$ Solve ILP in \eqref{formalOpt}-\eqref{max_edge} over $\hat{\mathcal{G}}$ instead of $\mathcal{G}$
\State Return $\gamma$
\end{algorithmic}
\label{BasicColGenAlg}
\end{algorithm}
%

\begin{align}
\label{formalOpt}
    \min_{\substack{\gamma_g  \in \{ 0,1\} \, \forall g \in \mathcal{G}}}
    \sum_{g \in \mathcal{G}}\Gamma_g \gamma_g\\
    \label{max_delv}
    \sum_{g \in \mathcal{G}} G_{dg}\gamma_g \leq 1\quad \forall d \in \mathcal{D} \\
    \label{max_time}
    \sum_{g \in \mathcal{G}} G_{tg}\gamma_g \leq N \quad \forall t \in \mathcal{T}  \\
    \label{max_rob}
    \sum_{g \in \mathcal{G}} G_{rg}\gamma_g = 1 \quad \forall r \in \mathcal{R} 
    \\
    \label{max_pos}
    \sum_{g \in \mathcal{G}} G_{pg}\gamma_g \leq 1 \quad \forall p \in \mathcal{P}  \\
        \label{max_edge}
    \sum_{g \in \mathcal{G}} G_{eg}\gamma_g \leq 1 \quad \forall e \in \mathcal{E}  
\end{align}

In \eqref{formalOpt}, we minimize the cost (that is, maximize the reward) of the MRP solution. In \eqref{max_delv}, we  enforce that no item is serviced more than once.  In \eqref{max_time}, we enforce that no more than the available number of robots $N$ is used at any given time.  In \eqref{max_rob}, we enforce that each extant robot is associated with exactly one route. In \eqref{max_pos}, we enforce that no more than one robot can occupy a given space-time position. In \eqref{max_edge}, we enforce that no more than one robot can move along any space-time edge.  

We describe a set of feasibility constraints and cost terms for robot routes in our application. 
 (1) Each item $d \in \mathcal{D}$ can only be picked up during its time window $[t^-_d,t^+_d]$. (2) Each item $d \in \mathcal{D}$ uses $c_d \in \mathbb{Z}_{+}$ units of capacity of a robot.  The capacity of a robot is $c_0 \in \mathbb{Z}_{+}$.  An  active robot $r \in \mathcal{R}$ is associated with an initial space-time position $p_{0r}$ (at the initial time, i.e., time 1)  and a remaining capacity $c_r \in [0, c_0]$.  

The cost associated with a robot route is defined by the following terms.  (1) $\theta_d \in \mathbb{R}_{-}$ is the cost associated with servicing item $d$.  (2) $\theta_1,\theta_2 \in \mathbb{R}_{0+}$ are the costs of being on the floor and moving respectively, which depreciate the robot.  Using $\theta_d$, $\theta_1$, and $\theta_2$,  we write $\Gamma_g$ as follows. $\Gamma_{g}=\sum_{d \in \mathcal{D}}G_{dg}\theta_d+\sum_{t \in \mathcal{T}}\theta_1G_{tg}+\sum_{e \in \mathcal{E}}\theta_{2}G_{eg}$
%
%
\section{Column Generation for MRP}
\label{CGSec}
Since in practice $\mathcal{G}$ cannot be enumerated, we attack optimization in  \eqref{formalOpt}-\eqref{max_edge} using column generation (CG). Specifically, we relax $\gamma$ to be non-negative and construct a sufficient set $\hat{\mathcal{G}} \subset \mathcal{G}$ to solve optimization over $\mathcal{G}$ using CG. CG iterates between solving the LP relaxation of \eqref{formalOpt}-\eqref{max_edge} over $\hat{\mathcal{G}}$, which is referred to as the Restricted Master Problem (RMP), followed by adding elements to $\hat{\mathcal{G}}$ that have negative reduced cost, which is referred to as pricing.  Below we formulate pricing as an optimization problem using $\lambda_{d}$, $\lambda_t$, $\lambda_r$, $\lambda_p$, and $\lambda_e$ to refer to the dual variables over constraints \eqref{max_delv}-\eqref{max_edge} of the RMP respectively.    
\begin{align}
\label{pricing_prob}
    \min_{g \in \mathcal{G}}\bar{\Gamma}_g \quad \mbox{where} \quad 
    \bar{\Gamma}_g=\Gamma_g -\sum_{i \in \mathcal{I}} \lambda_i G_{ig}  
\end{align}
We terminate optimization when the solution to \eqref{pricing_prob} is non-negative, which means that $\hat{\mathcal{G}}$ is provably sufficient to exactly solve the LP relaxation of optimization over $\mathcal{G}$~\citep{lubbecke2005selected}.  We initialize $\hat{\mathcal{G}}$ with any feasible  solution (perhaps greedily constructed) so as to ensure that each $r \in \mathcal{R}$ is associated with a route.  At termination of CG, if $\gamma_g \in \{0,1\}, \forall g \in \mathcal{G}$, then the solution, i.e. the tracks defined by $\{g\in \mathcal{G} | \gamma_g=1\}$, is provably optimal.  Otherwise, an approximate solution can be produced by solving the ILP formulation over $\hat{\mathcal{G}}$ or the formulation can be tightened using valid inequalities, such as subset row inequalities \citep{jepsen2008subset}.  We can also use branch-and-price \citep{barnprice} to formulate CG inside a branch-and-bound formulation.  Algorithm~\ref{BasicColGenAlg} shows pseudocode for CG.  We show an enhanced version of CG motivated by dual optimal inequalities (DOI) that we propose in Appendix \ref{doi_sec}
%
%
\section{Solving the Pricing Problem }
\label{Pricing}
In this section, we consider  the problem of pricing, which we show is a resource-constrained shortest-path problem (RCSP) \citep{righini2008new}.  We organize this section as follows.  In Section \ref{trivialPricing}, we  formulate pricing as an RCSP over a graph whose nodes correspond to space-time positions and whose resources correspond to the items picked up. In Section \ref{efficentPricing}, we accelerate computation from Section \ref{trivialPricing} by coarsening the graph, leaving only locations of significance such as item locations across time. In Section \ref{removeExplicitTIme}, we further accelerate computation by limiting the times considered while still achieving exact optimization during pricing.  In Section \ref{partOptFaster}, we show that CG can be accelerated by updating the $\lambda_i$ for all $i \in \mathcal{D}\cup \mathcal{R}$ more often than the remainder of the dual solution, saving computation time by precluding the need to reconstruct the coarsened graph as often between rounds of pricing. 
\subsection{Basic Pricing}
\label{trivialPricing}
In this section we establish a weighted graph admitting an injunction from the routes in $\mathcal{G}$ to the paths in the graph.  For a given route $g$, the sum of the weights along the corresponding path in the weighted graph is equal to the route's reduced cost $\bar{\Gamma}_g$.  Thus finding the lowest-cost feasible path in this graph solves Eq \eqref{pricing_prob}.  The graph proposed is a modified form of the time-extended graph $(\mathcal{P},\mathcal{E})$.  Nodes are added to represent start/end locations, item pickups, and the use of an extant robot.  Weights are amended by the corresponding dual variables associated with a given node/edge.  We solve a RCSP over this graph where the resources are the items to be pick up. 

Formally, consider a graph $(\mathcal{P}^+,\mathcal{E}^+)$ with paths described by $x_{p_ip_jg}\in \{0,1\}$ for $(p_i,p_j) \in \mathcal{E}^+,g \in \mathcal{G}$, where $x_{p_ip_jg}=1$ indicates that edge $(p_i,p_j)$ is traversed by the path on the graph corresponding to route $g$. Each edge $(p_i,p_j)$ has an associated weight $\kappa_{p_ip_j}$.  There is a node in $\mathcal{P}^+$ for each $p\in \mathcal{P}$, for each pairing of $d\in \mathcal{D}$ and $t \in [t_d^-,t_d^+]$ denoted $p_{dt}$, for each $r \in \mathcal{R}$ denoted $p_r$, the source node $p_+$, and the sink node $p_-$.  We ensure that  $\bar{\Gamma}_g=\sum_{(p_i,p_j) \in \mathcal{E}^+}\kappa_{p_ip_j}x_{p_ip_jg}$ for all $g \in \mathcal{G}$.  For each pair of space-time positions $p_i,p_j$ occurring at the same cell at times $t_i,t_j=t_i+1$ (representing a wait action),  we set  $\kappa_{p_ip_j}=\theta_{1}-\lambda_{t_j}-\lambda_{p_j}$.  We set $x_{p_ip_jg}=1$ IFF robot route $g$ transfers from $p_i$ to $p_j$ and no pickup is made at $p_i$.

For each pair of space-time positions $p_i,p_j$ occurring at times $t_i$ and $t_j=t_i+1$ and associated with space-time edge $e$ (representing a move action), we set $\kappa_{p_ip_j}=\theta_{1}+\theta_{2}-\lambda_e-\lambda_{t_j}-\lambda_{p_j}$. We set $x_{p_ip_jg}=1$ IFF robot route $g$ transfers from $p_i$ to $p_j$ and no pickup is made at $p_i$.
For each $d \in \mathcal{D},t\in [t_d^-,t_d^+]$, which occurs at space-time position $p$, we set $\kappa_{p p_{dt}}=\theta_d-\lambda_{d}$.  We set $x_{p p_{dt}g}=1$ IFF robot route $g$  picks up item $d$ at time $t$.  For each $d \in \mathcal{D},t\in [t_d^-,t_d^+]$, which occurs at an associated $p$, we provide identical outgoing $\kappa$ terms for $p_{dt}$ as we do $p$ (except there is no self connection $p_{dt}$ to $p_{dt}$).  We set $x_{p_{dt}p_jg}=1$ IFF robot route $g$ transfers from the position of item $d$ to $p_j$ and item $d$ is picked up at time $t_j-1$ on route $g$. 
For each $t \in \mathcal{T}$ we connect the $p_+$ to the launcher at time $t$ denoted $p_{0t}$ with weight $ \kappa_{p_+p_{0t}}=\theta_1-\lambda_{t}-\lambda_{p_{0t}}$.  We set $x_{p_+p_{0t}g}=1$ IFF the robot route $g$ appears first at  $p_{0t}$.  
For each $r \in \mathcal{R}$  we set $\kappa_{p_+p_{r}}=\theta_1-\lambda_r-\lambda_{t=1}-\lambda_{p_r} $. We set $x_{p_+p_{r}g}=1$ IFF the robot route $g$ appears first at $p_{r}$.  For each $r \in \mathcal{R}$, $p_r$ has one single outgoing connection to $p_{0r}$ with weight $\kappa_{p_rp_{0r}}=0$.

For each $t \in \mathcal{T}$ we set $\kappa_{p_{0t}p_-}=0$.  We set $x_{p_{0t}p_-g}=1$ IFF the robot route $g$ has its last position at  $p_{0t}$.
Using $\kappa$ defined above we express the solution to \eqref{pricing_prob} as an ILP ( followed by description) using decision variables $x_{p_ip_j}\in \{0,1\}$ where $x_{p_ip_j}$ is equal to $x_{p_ip_jg}$ for all $(p_i,p_j) \in \mathcal{E}^+$.
%
\begin{align}
\label{objPath}
    \min_{x_{p_ip_j} \in \{0,1\} \quad \forall (p_i,p_j) \in \mathcal{E}^+}\sum_{(p_i,p_j) \in \mathcal{E}^+}\kappa_{p_ip_j}x_{p_ip_j}\\
    \label{flowConst}
    \sum_{(p_i,p_j) \in \mathcal{E}^+} x_{p_ip_j}-\sum_{(p_j,p_i) \in \mathcal{E}^+} x_{p_jp_i}=[p_i=p_+]-[p_i=p_-] \quad \forall p_i \in \mathcal{P}^+  \\
    \label{capacityConst}
    \sum_{d \in \mathcal{D}} c_d \sum_{t_{d}^-\leq t\leq t_{d}^+} \sum_{(p,p_{dt}) \in \mathcal{E}^+} x_{pp_{dt}} \leq c_0+\sum_{r \in \mathcal{R}}(c_r-c_0)x_{p_+p_r}  \\
    \label{resourceConst}
    \sum_{t_{d}^-\leq t\leq t_{d}^+}  \sum_{(p,p_{dt}) \in \mathcal{E}^+} x_{pp_{dt}} \leq 1 \quad \forall d \in \mathcal{D}  
\end{align}
In \eqref{objPath} we provide objective s.t. $\bar{\Gamma}_g=\sum_{(p_i,p_j) \in \mathcal{E}^+}\kappa_{p_ip_j}x_{p_ip_jg}$ for all $g \in \mathcal{G}$.  In \eqref{flowConst} we ensure that $x$ describes a path from $p_+$ to $p_-$ across space time. In \eqref{capacityConst} we ensure that capacity is obeyed. In \eqref{resourceConst} we ensure that each item is picked up at most once.  Optimization in \eqref{objPath}-\eqref{resourceConst} is strongly NP-hard as complexity grows exponentially with $|\mathcal{D}|$ \citep{Desrochers1992}.  

\subsection{Efficient Pricing:  Considering Only Nodes Corresponding to Items }
\label{efficentPricing}
In this section we rewrite the optimization for pricing in a manner that vastly decreases graph size allowing optimization to be efficiently achieved for the RCSP solver.  We exploit the fact that given the space-time positions where item pickups occur, we can solve of the remainder of the problem as independent parts.  Each such independent part is solved as a shortest path problem, which can be solved via a shortest path algorithm such as Dijkstra's algorithm \citep{dijkstra1959note}. 

We now consider a graph with node set $\mathcal{P}^2$  with edge set $\mathcal{E}^2$, decision $x^2_{p_ip_jg} \in \{0,1\}$ and weights $\kappa^2$.  There is one node in $\mathcal{P}^2$ for each $p \in \mathcal{P}^+$ excluding those for $p \in \mathcal{P}$, i.e., $\mathcal{P}^2 = \mathcal{P}^+\setminus\mathcal{P}$.  
For any $p_i,p_j \in \mathcal{P}^2$, $(p_i,p_j) \in \mathcal{E}^2$ IFF there exists a path from $p_i$ to $p_j$ in $\mathcal{E}^+$ traversing only intermediate nodes that exist in $\mathcal{P}$. We set $\kappa^2_{p_ip_j}$ to be the weight of the shortest path from $p_i$ to $p_j$ in $\mathcal{E}^+$ using only intermediate nodes in $\mathcal{P}$.  This is easily computed using a shortest path algorithm.  We set $x^2_{p_ip_jg}=1$ IFF  $p_i$ is followed by $p_j$ in robot route $g$ when ignoring nodes in $\mathcal{P}$.  Replacing $\mathcal{E}^+,x$ with $\mathcal{E}^2,x^2$ respectively in \eqref{objPath}-\eqref{resourceConst} we have a smaller but equivalent optimization problem permitting more efficient optimization.


%
\subsection{ More Efficient Pricing: Avoiding Explicit Consideration of All Times}
\label{removeExplicitTIme}
The optimization in Eq \eqref{objPath}-\eqref{resourceConst} over $\mathcal{E}^2$ requires the enumeration of all $d \in \mathcal{D}, t \in [t_d^-,t_d^+]$, which is expensive.  In this section we circumvent the enumeration of all $d \in \mathcal{D}, t \in [t_d^-,t_d^+]$ pairs by aggregating time into sets in such a manner so as to ensure exact optimization during pricing.  
For every $d \in \mathcal{D}$, we construct $\mathcal{T}_{d}$, which is an ordered subset of the times $[t_d^-,t_d^+ +1]$ where $\mathcal{T}_d$ includes initially $t_d^-$ and $t_d^+ +1$ and is augmented as needed.  We order these in time where $\mathcal{T}_{dj}$ is the $j$'th value ordered from earliest to latest.  $\mathcal{T}_d$ defines a partition of the window $[t_d^-,t_d^+]$ into $|\mathcal{T}_d|-1$ sets, where the $j$'th set is defined by $[\mathcal{T}_{dj},\mathcal{T}_{dj+1}-1]$

We use $\mathcal{P}^3,\mathcal{E}^3,\kappa^3,x^3$ to define the graph and solution mapping.  Here $\mathcal{P}^3$ consists of $p_+,p_-, p_r \forall r \in \mathcal{R}$ and one node $p_{dj}$ for each $d \in \mathcal{D},j \in \mathcal{T}_{d}$.  We define $x^3_{p_+p_{dj}g}=1$ if route $g$ services item $d$ at a time in $[\mathcal{T}_{dj},\mathcal{T}_{d \; j+1}-1]$ as its first pick up.  The remaining $x$ terms are defined similarly over aggregated time sets.  We assign each $\kappa^3_{p_ip_k}$ to be some minimum $\kappa^2$ over the possible paths in $(\mathcal{P}^2,\mathcal{E}^2)$ associated with $p_i,p_k\in \mathcal{P}^3$.  We set $\kappa^3_{p p_{dj}}=\min_{t \in [\mathcal{T}_{dj},\mathcal{T}_{d \; j+1}-1]} \kappa^2_{p p_{dt}}$ for all $p \in \{p_+,p_r \forall r \in \mathcal{R}\}$.  We set $\kappa^3_{ p_{+}p_{r}} = \kappa_{ p_{+}p_{r}}$.  We set $\kappa^3_{ p_{dj}p_-}=\min_{t \in [\mathcal{T}_{dj},\mathcal{T}_{d \; j+1}-1]} \kappa^2_{p_{dt}p_-}$.  For any pair of unique $d_i,d_k$ and windows $j_i,j_k$ we set  $\kappa^3_{ p_{d_ij_i}p_{d_kj_k}}=\min_{\substack{t_0 \in [\mathcal{T}_{d_ij_i},\mathcal{T}_{d_i \; j_i+1}-1]\\ t_1 \in [\mathcal{T}_{d_kj_k},\mathcal{T}_{d_k \; j_k+1}-1]}} \kappa^2_{p_{d_it_0}p_{d_kt_1}}$.  Evaluating each of the $\kappa^3$ terms amounts to solving a basic shortest path problem (no resource constraints), meaning not all $\kappa^2$ terms mentioned in these optimizations need be explicitly computed.  
Replacing $\mathcal{E}^+$ with $\mathcal{E}^3$ in \eqref{objPath}-\eqref{resourceConst} we have a smaller optimization problem permitting more efficient optimization, which provides a lower bound on   \eqref{objPath}-\eqref{resourceConst}.  

Optimization produces a feasible route when each item in the route is associated with exactly one unique time.  In pursuit of a feasible route, we add the times associated with items in the route to their respective $\mathcal{T}_d$ sets.  We iterate between solving the RCSP over $\mathcal{E}^3$ and augmenting the $\mathcal{T}_d$ until we obtain a feasible route.  This must ultimately occur since eventually $\mathcal{T}_d$ would include all $t \in \mathcal{T}$ for all $d \in \mathcal{D}$.  Though it should occur much earlier in practice.  We provide pseudocode for this pricing method in Algorithm \ref{fastPricing} in Appendix \ref{appendix:a}.

%

\subsection{Partial Optimization of the Restricted Master Problem for Faster pricing}
\label{partOptFaster}
Solving the pricing problem is the key bottleneck in computation experimentally.  One key time consumer in pricing is the computation of the $\kappa$ terms, which can easily be avoided by observing that $\kappa^2,\kappa^3$ terms are offset by changes in $\lambda_d$ and $\lambda_r$ but the actual route does not change so long as $\lambda_e$, $\lambda_p$, and $\lambda_t$ are fixed. We resolve the RMP fully only periodically so that we can perform several round of pricing using different $\lambda_d,\lambda_r$ terms leaving the $\lambda_e,\lambda_p,\lambda_t$ fixed. 

%

\section{Experiments}
\label{exper}

We run two sets of experiments to empirically study our model.  In the first set, we test our model on two classes of random, synthetic problem instances, recording relevant performance and solution statistics.  We take a close look at the distribution of these results.  Next we compare our algorithm to a modified version employing MAPF to assess the added value of the algorithm's consideration of robot collisions in the formulation.

\subsection{Algorithm Performance}

We study the performance of our algorithm on two distinct problem classes where each class includes a set of 100 random instances with specific, shared parameters.  Each class shares the same grid size, number of time steps, number of serviceable items, number of map obstacles, and number of robots.  Given a set of problem parameters, a single instance additionally includes a random set of obstacle locations, item locations and their respective demands and time windows, and extant robot start locations.  We solve each instance over the class, recording the LP objective solution and solving the corresponding ILP over the column set $\hat{\mathcal{G}}$ obtained through CG.  For each class of problems, to establish the algorithm's performance and the quality of its solutions, we look at the distribution of the times and numbers of iterations required for CG to converge, the LP objective of the CG solution, and the corresponding relative gaps.  The relative gap is defined as the the absolute difference between our integer solution (the upper bound) and the lower bound (the LP objective solution) divided  by the lower bound.  We normalize so as to efficiently compare the gap obtained (upper bound - lower bound) across varying problem instances.  Experiments are run in MATLAB and CPLEX is used as our general purpose MIP solver.

We solve the RCSP in pricing using an exponential time dynamic program outlined in Appendix \ref{appendix:c}.  In each round of pricing we return the twenty lowest reduced cost columns we obtain, if they all have negative reduced cost.  Otherwise, we return as many negative reduced cost columns as we obtain.  We update $\lambda_t$, $\lambda_p$, $\lambda_e$, and the associated graph components every three CG iterations, unless we are unable to find a negative reduced cost column in a given iteration, in which case update all dual variables and rerun pricing.  If at any point pricing fails to find a negative reduced cost column while all dual variables are up to date, then we have finished optimization and we conclude CG.  To ensure feasibility for the initial round of CG, we initialize the RMP with a prohibitively high cost dummy route $g_{r,init}$ for each $r \in \mathcal{R}$, where all $G_{dg_{r,init}},G_{tg_{r,init}},G_{pg_{r,init}},G_{eg_{r,init}}=0$ but $G_{rg_{r,init}}=1$.  These dummy routes represent and active robot route and thus guarantee that Eq \ref{max_rob} is satisfied.  They ensure feasibility, but are not active at termination of CG due to their prohibitively high cost.

\begin{figure}[!hbtp]
    \centering
	\includegraphics[width=0.25\linewidth]{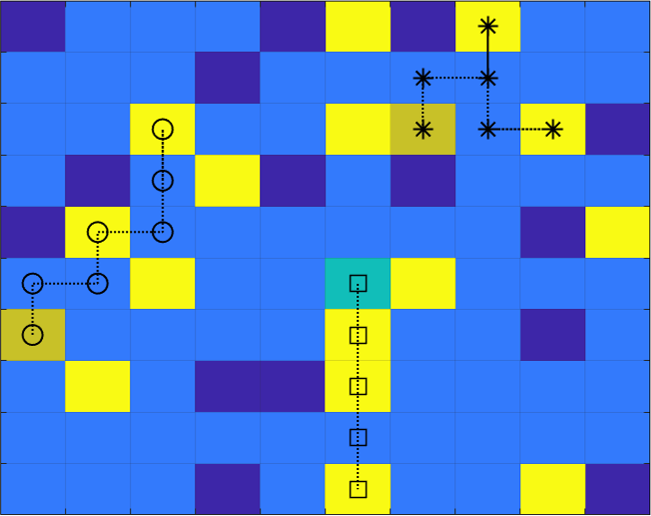}
	\includegraphics[width=0.25\linewidth]{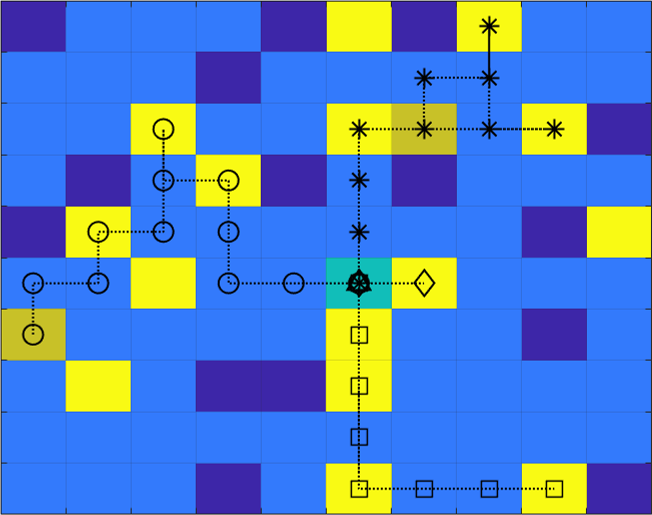}
    \includegraphics[width=0.25\linewidth]{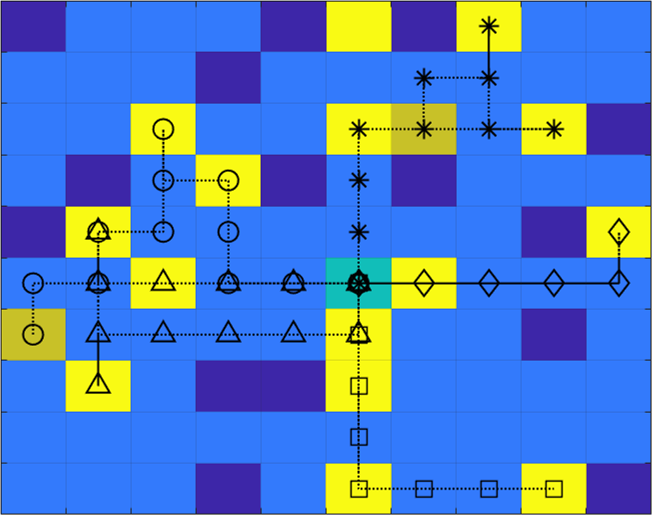} \\
    \includegraphics[width=0.50\linewidth]{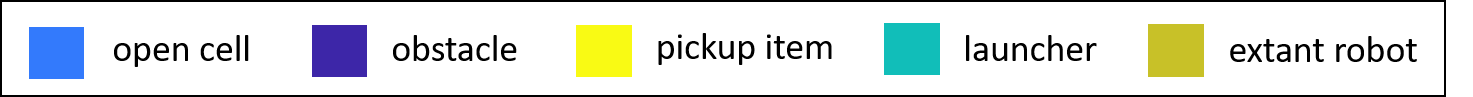}
	\caption{Robot route results for a single instance over 3 snapshots in time.  Each track is a robot route up through that time step.  Traversable cells, obstacles, the starting/ending launcher, item locations, and extant robot locations are all noted in the legend.
		\textbf{(Left): t = 8 snapshot}   
		\textbf{(Middle): t = 16 snapshot}
		\textbf{(Right): t = 30 (end time) snapshot}}
	\label{fig:paths}
\end{figure}

In our first class of problems we use a 10x10 grid, 4 total robots with 2 initially active, 15 serviceable items, and 30 total time steps.  Each robot, including the extant ones, has a capacity of 6, while each item has a random capacity consumption uniformly distributed over the set \{1,2,3\}.  We set both $\theta_{1}$ and $\theta_{2}$ to 1, and the reward for servicing any item, $\theta_{d}$, is -50.  Each item's time window is randomly set uniformly over the available times and can be up to 20 time periods wide.  Each map has 15 random locations chosen to serve as obstacles that are not traversable.  We solve 100 unique random instances and aggregate the results in Table \ref{table:runtime_10}.  A sample problem with the solution routes is shown in Figure \ref{fig:paths}.  Each plot in the Figure \ref{fig:paths} shows a snapshot in time of the same instance's solution.  A snapshot shows each robot's route from the initial time up to the time of the snapshot.

\begin{table*}[t!]
	\centering
	\scalebox{0.8}{
		\begin{tabular}{|c|c|c|c|c|c|} 
		\hline
		  & \bf Time (sec) & \bf Iterations & \bf LP Objective & \bf Integral Objective & \bf Relative Gap\\
		\hline
		\bf mean & 236.3 & 24.7 & -581.0 & -574.6 & .01   \\
		\hline
		\bf median & 160.0 & 24 & -586.2 & -581 & .01  \\
		\hline
	\end{tabular}}
	\caption{10x10 grid results over 100 random problem instances}
	\label{table:runtime_10}
\end{table*}

We see that over the problem instances in this class we have an average run time of 160 seconds requiring on average 24.7 CG iterations.  We report an average LP objective of -581.0 and an average relative gap of .01, thus certifying that we are efficiently producing near optimal solutions.  In 93 out of the 100 instances the approximate solution to Eq \eqref{formalOpt}-\eqref{max_edge} reused robots after they returned to the launcher.

In our second class of problems we use a 20x20 grid, 5 total robots with 2 initially active, 25 serviceable items, and 100 total time steps.  Each robot, including the extant ones, has a capacity of 6 while each item has a random capacity consumption uniformly distributed over the set \{1,2,3\}.  We set both $\theta_{1}$ and $\theta_{2}$ to 1 and the reward for servicing any item, $\theta_{d}$, is -50.  Each item's time window is  randomly set uniformly over the available times and can be up to 8 time periods wide.  Each map has 40 random locations chosen to serve as obstacles that are not traversable.  We run on 100 unique random instances and aggregate the results in Table \ref{table:runtime_20}.

\begin{table*}[t!]
	\centering
	\scalebox{0.8}{
		\begin{tabular}{|c|c|c|c|c|c|} 
		\hline
		\bf  & \bf Time (sec) & \bf Iterations & \bf LP Objective & \bf Integral Objective & \bf Relative Gap\\
		\hline
		\bf mean & 478.8 & 30.1 & -639.4 & -6329.5 & .02 \\
		\hline
		\bf median & 389.6 & 30.0 & -643.5 & -632 & .01 \\
		\hline
	\end{tabular}}
	\caption{20x20 grid results over 100 random problem instances}
	\label{table:runtime_20}
\end{table*}

We see in this class of instances we get an average run time of 478.8 seconds and an average of 30.1 iterations of CG.  We get an average LP objective of -639.4 and a relativity gap of .01, again showing that we are efficiently producing near optimal solutions.  In all 100 instances the approximate solution to Eq \eqref{formalOpt}-\eqref{max_edge} reused robots after they returned to the launcher.

We see a slight increase in the iterations required for the second problem class with respect to the first problem class.  We see a larger growth in the time required for convergence.  We expect this trend can be alleviated with the application of heuristic pricing \citep{danna2005branch,FlexDOIArticle}.  Heuristic pricing attacks the pricing optimization problem through the use of heuristic methods.  Since we need only produce a negative reduced cost route through each round of pricing, not necessarily the minimum one, heuristic pricing can hasten CG by saving computational time.  Such a heuristic would produce approximate solutions with respect to the ordering of the items but still be optimal given a particular ordering.  We also see a very small increase in the relative gap on larger problem instances.  Though most problems on the 20x20 grid still have a very small gap, we start to see more problems with a gap close to 5\%.  The relative gap can be reduced by tightening the relaxation through the use of subset row inequalities \citep{jepsen2008subset, wang2017learning}.

\subsection{Comparison with MAPF}

We compare our algorithm to a modified version that incorporates MAPF.  This version will initially ignore robot collision constraints but ultimately consider them after a set of serviceable items are assigned to specific robots.  The modified algorithm works as follows.  We solve a given problem instance using our CG algorithm, but we neglect the collision constraints, meaning $\lambda_p=0,\lambda_e=0, \forall p\in \mathcal{P}, e\in \mathcal{E}$.  This closely resembles a vehicle routing problem \citep{Desrochers1992} and delivers us a set of robot routes, including the items serviced by each robot, however this could include collisions.  We then take the disjoint set of items serviced and feed them to a MAPF solver \citep{LiAAMAS20b}.  The MAPF solver delivers a set of non-colliding robot routes, each attempting to service the set of items assigned to it.  If the MAPF solver fails to provide a valid route for a particular robot (i.e., it cannot make it back to the launcher in time) that route is neglected in the algorithm's final solution.

\begin{figure}[!htbp]
    \centering
	\includegraphics[width=0.5\linewidth]{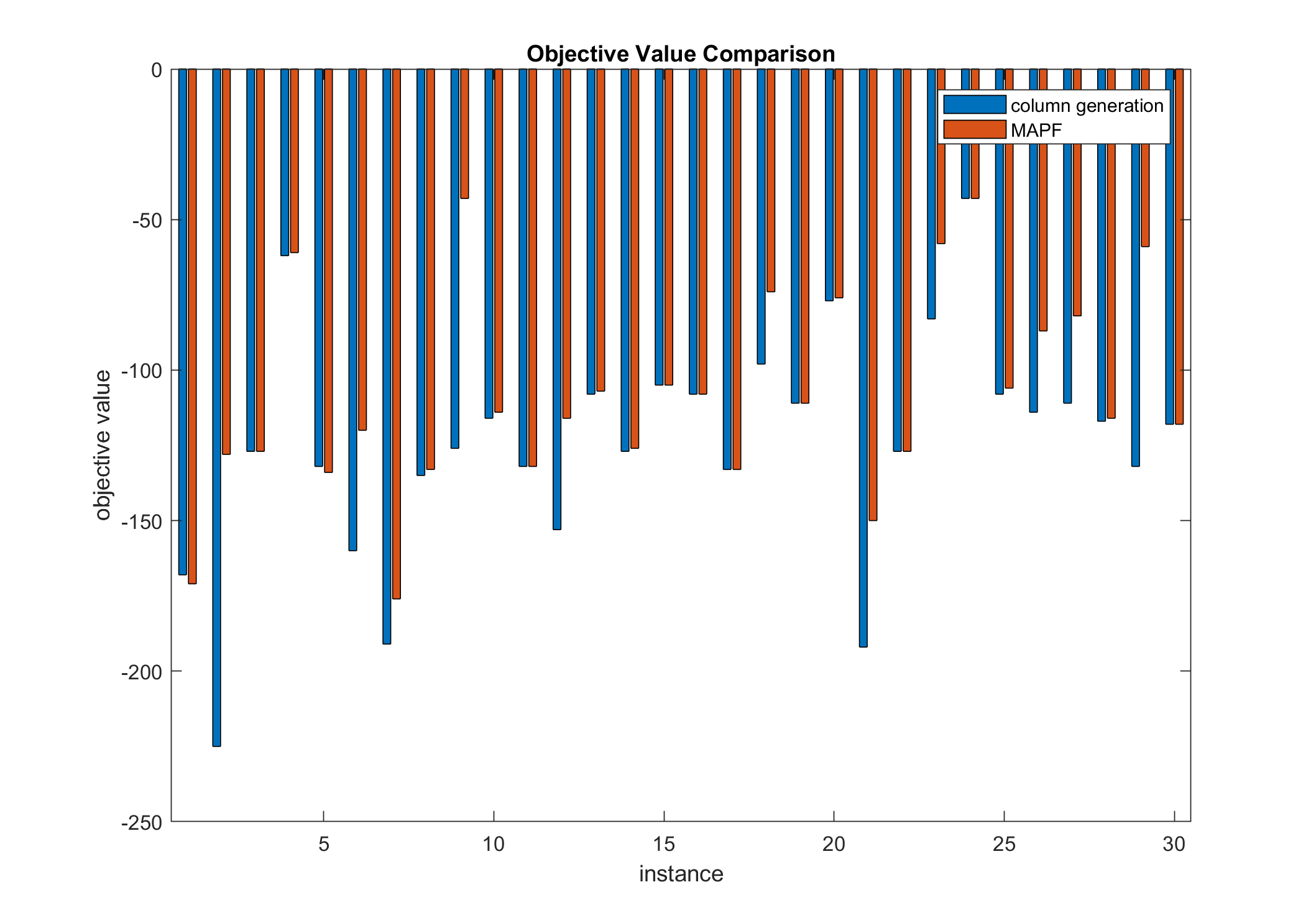}
	
	\caption{Objective values for both approaches over each problem instance.  Our full CG approach is shown in blue.  It is compared against the modified column CG + MAPF approach shown in orange.}
	\label{fig:mapf_comparison}
\end{figure}

We compare the resulting objective values from our full CG approach to this modified approach.  For the purposes of this comparison, we neglect time constraints for the items so as to be generous to the MAPF solver, which is not equipped to handle time windows for items.  We solve 30 random instances with the same parameters.  We use a 20x20 grid, 35 serviceable items, 100 random obstacles, 9 total robots, 1 extant robot, and 25 total time steps.  We set $\theta_{1}$ to 1, $\theta_{2}$ to 0, and the reward for servicing any item, $\theta_{d}$, to -15.  The objective value results for both approaches are show in table \ref{table:mapf_comparison}. A side by side plot of the objective values are shown in Figure \ref{fig:mapf_comparison}.

\begin{table*}[t!]
	\centering
	\scalebox{0.8}{
		\begin{tabular}{|c|c|c|c|} 
		\hline
		    & \bf CG & \bf modified CG + MAPF & \bf Difference (CG - MAPF) \\
		\hline
		\bf mean & -124.6 & -116.8 & -7.9 \\
		\hline
		\bf median & -122.0 & -111.0 & -1.0 \\
		\hline
	\end{tabular}}
	\caption{Objective value results for both algorithms over 30 random instances.  Our full approach is labeled CG.  We compare against modified CG + MAPF.}
	\label{table:mapf_comparison}
\end{table*}

We see an average objective difference of -7.9 and a median difference of -1.0 from the modified algorithm to our full algorithm.  We note from looking at Figure \ref{fig:mapf_comparison} that many instances deliver very similar objective results, however some show drastic improvements for our algorithm.  These instances largely include robot routes that the MAPF algorithm was unable to find a complete route for within the time constraint given the potential collisions with other robots.  With such problems we see it is critical to employ our full algorithm that jointly considers routing and assignment.


\section{Conclusions}
\label{conc}
In this paper, we unified the work on multi-agent path finding with the vehicle routing/column generation literature to produce a novel approach applicable to broad classes of multi-robot planning (MRP) problems.  Our work treats MRP as a weighted set packing problem where sets correspond to valid robot routes and elements correspond to space-time positions.  Pricing is treated as a resource-constrained shortest-path problem (RCSP), which is NP-hard but solvable in practice \citep{irnich2005shortest}.  We solve the RCSP by adapting the approach of \citep{boland2017continuous} to limit the time windows that need be explored during pricing.

In future work, we seek to tighten the LP relaxation using subset row inequalities \citep{jepsen2008subset} and ensure integrality with branch-and-price \citep{barnprice}. Subset row inequalities are trivially applied to sets over the pickup items since they do not alter the solution paths.  Similarly, branch-and-price would be applied following the vehicle routing literature to sets over pickup items \citep{Desrochers1992}.  As well, we intend to incorporate heuristic pricing to solve the resource-constrained shortest-path problem in pricing more efficiently, thus increasing the scalability of the algorithm.  We also seek to provide insight into the structure of dual optimal solutions and study the effect of smoothing in the dual, based on the ideas of \citep{haghani2020smooth,haghani2020relaxed}.  Simply put, we suspect that dual values should change smoothly across space and time, thus we will encourage such solutions over the course of column generation.

\bibliographystyle{abbrvnat} 
\bibliography{col_gen_bib}

\appendix

\section{More Efficient Pricing Algorithm}
\label{appendix:a}

In this section, we provide pseudocode for the pricing method described in Section \ref{removeExplicitTIme} as an algorithm.  We use $t_{p_{d_ij_i}p_{d_kj_k}0}$ and $t_{p_{d_ij_i}p_{d_kj_k}1}$ to denote the minimizers $ \argmin_{\substack{t_0 \in [\mathcal{T}_{d_ij_i},\mathcal{T}_{d_i \; j_i+1}-1]\\ t_1 \in [\mathcal{T}_{d_kj_k},\mathcal{T}_{d_k \; j_k+1}-1]}} \kappa^2_{p_{d_it_0}p_{d_kt_1}}$ used to calculate $\kappa^3_{ p_{d_ij_i}p_{d_kj_k}}$.  The term $t_{p_{d_ij_i}p_{d_kj_k}0}$ is the time component minimizer for $p_{d_ij_i}$ while $t_{p_{d_ij_i}p_{d_kj_k}}$ is the time component minimizer for $p_{d_kj_k}$.  These are the outgoing and incoming times, respectively, for the shortest path on $(\mathcal{P}^2,\mathcal{E}^2)$ between $p_{d_ij_i}$ and $p_{d_kj_k}$.  We use tot\_sz to keep track of the total number of elements in all $\mathcal{T}_d$ sets.  A growth in tot\_sz implies a potential mismatch between the incoming time and the outgoing time at an item location.  In such cases, tot\_sz grows to narrow the time ranges for the sets, making it less likely to have a mismatch.  When tot\_sz does not grow, no mismatch is possible and the solution obtained is guaranteed to represent a feasible route, therefore we terminate pricing.  Algorithm \ref{fastPricing} shows pseudocode for the pricing method described in Section \ref{removeExplicitTIme}.

\begin{algorithm}[!h]
 \caption{Fast Pricing
 }

\begin{algorithmic}[!h] 
\State $\mathcal{T}_d \leftarrow [t_d^-,t_d^++1] \quad \forall d \in \mathcal{D}$
\label{init_sets}
\Repeat
\label{bigLoop}
\State tot\_size $\leftarrow \sum_{d \in \mathcal{D}}|\mathcal{T}_{d}|$
\label{getSz}
\State   $x\leftarrow $ Solve  Eq \eqref{objPath}-\eqref{resourceConst} over $\mathcal{E}^3$
\For{($p_{i},p_{k}) \in  \mathcal{E}^3$ s.t. $ x_{p_i,p_k}=1$ }
\If {$p_i \neq p_+$ and $p_i \neq p_r  $  for any $r \in \mathcal{R}$} 
\State Let $p_i$ correspond to item $d$
\State $\mathcal{T}_d \leftarrow \mathcal{T}_d \cup t_{p_i,p_k,0} $
\EndIf
\If {$p_k \neq p_-$ and $p_k \neq p_r  $  for any $r \in \mathcal{R}$} 
\State Let $p_k$ correspond to item $d$
\State $\mathcal{T}_d \leftarrow \mathcal{T}_d \cup t_{p_i,p_k,1} $
\EndIf
\EndFor
 \Until{ tot\_size=$\sum_{d \in \mathcal{D}}|\mathcal{T}_{d}|$}
\State Let $g$ correspond to the solution to \eqref{objPath} computed via optimization over $\mathcal{E}^3$
\State Return $g$
\end{algorithmic}
\label{fastPricing}
\end{algorithm}

\section{Dual Optimal Inequalities}
\label{doi_sec}
In this section we provide dual optimal inequalities (DOI) for MRP, which accelerate CG and motivates better approximate solutions at termination of CG when the LP relaxation is loose.  Our DOI are motivated by the following observation.  No optimal solution to \eqref{pricing_prob} services an item $d$ that is associated with a net penalty instead of a net reward for being serviced, meaning that $\theta_d\leq \lambda_d $ must be observed. 
This is because simply not servicing the item but using an identical route in space time would produce a lower reduced cost route.  Since the dual LP relaxation of  \eqref{formalOpt}-\eqref{max_edge} is increasing with respect to $\lambda$ no optimal dual solution to  \eqref{formalOpt}-\eqref{max_edge} will violate the following $\theta_d\leq \lambda_d \quad  \forall d \in \mathcal{D}$.
By enforcing $\theta_d\leq \lambda_d \quad  \forall d \in \mathcal{D}$ at each iteration of CG optimization we accelerate CG by restricting the dual space that need be explored.  In the primal form, Eq \eqref{formalOpt} and Eq \eqref{max_delv} are altered as follows with primal variables $\xi_d$ corresponding to $\theta_d\leq \lambda_d \quad  \forall d \in \mathcal{D}$.

\begin{align}
   \eqref{formalOpt}\mbox{ becomes } \min_{\substack{\gamma_g\geq 0 \\ \xi\geq 0}}
    \sum_{g \in \mathcal{G}}\Gamma_g \gamma_g -\sum_{d \in \mathcal{D}}\theta_d \xi_d \nonumber\\
    \mbox{and }\eqref{max_delv}\mbox{ becomes }\sum_{g \in \mathcal{G}} G_{dg}\gamma_g \leq 1+\xi_d\quad \forall d \in \mathcal{D} \nonumber
\end{align}

In our experiments we only use the replacements above when solving the ILP over the column set $\hat{\mathcal{G}}$.  When enforcing that $\gamma$ is binary, the technique described often leads to a closer approximations to the solution to Eq \eqref{formalOpt}-\eqref{max_edge}.  We map any solution derived this way to one solving the original ILP by arbitrarily removing over-included items from routes in the outputted solution until each item is included no more than once.

\section{Resource-Constrained Shortest-Path Solver}
\label{appendix:c}
We solve the resource-constrained shortest-path problem (RCSP) in pricing via an exponential time dynamic program that iterates over the possible remaining capacity levels for a robot (starting at the highest), enumerating all available routes corresponding to paths in $(\mathcal{P}^3,\mathcal{E}^3)$ at each capacity level, and then progressing to the down to the next highest remaining capacity level.  At each level we eliminate any inferior routes.  We call a route inferior to another if all of the following are satisfied: (1) it has the same remaining capacity and corresponding position in the node set $\mathcal{P}^3$ as the other, (2) it has lower cumulative edge cost on $(\mathcal{P}^3,\mathcal{E}^3)$ than the other, and (3) it has a set of serviceable items available to it that is a subset of the other's.

We start at the maximum robot capacity and enumerate all possible, single visit traversals.  We save a robot state for each such route.  A robot state is defined by its current corresponding position in the node set $\mathcal{P}^3$, the items serviced, the cost incurred so far on $(\mathcal{P}^3,\mathcal{E}^3)$, and the remaining capacity.  We set $\mathcal{K}_{p,h}$ to be the cost of a path at graph position $p \in \mathcal{P}^3$ with path history $h$, a set of all previously visited graph positions.  We set $\mathcal{C}_{p,h}$ to be the remaining capacity available for a robot at corresponding graph position $p$ with history $h$.  For a robot route with initial visit at item $d$ at corresponding graph position $p_{dj}$ we have the following remaining capacity and cost.

\begin{align}
\label{dp:cost_update}
    \mathcal{K}_{p_{dj},\{p_+\}} = \kappa^3_{ p_+p_{dj}}\\
    \mathcal{C}_{p_{dj},\{p_+\}} = c_0-c_d
\end{align}

We then move on to the next highest remaining robot capacity level.  For each saved robot state at this remaining capacity, we enumerate all available single visit traversals (including back to the launcher) and save a state for each route generated.  An item is available to be visited if that item has not yet been visited in the route and visiting it would not exceed the remaining capacity.  For a robot traveling from corresponding graph position $p_{d_ij_i}$ with history $h$, to corresponding graph position $p_{d_kj_k}$, we have the following update for the cost and remaining capacity.

\begin{align}
\label{dp:cost_update2}
    \mathcal{K}_{p_{d_kj_k},h\cup p_{d_ij_i} } = \mathcal{K}_{p_{d_ij_i},h} + \kappa^3_{ p_{d_ij_i}p_{d_kj_k}}\\
    \mathcal{C}_{p_{d_kj_k},h\cup p_{d_ij_i}} = \mathcal{C}_{p_{d_ij_i},h} - c_{d_k}
\end{align}

We eliminate all inferior routes generated and continue on to the next capacity level until we have exhausted all possible remaining capacity levels.  At the end we have series of routes drawn out, including the route with minimum cost on $(\mathcal{P}^3,\mathcal{E}^3)$.  We can return any number of these that have a negative cost.  Returning more serves to reduce the number of CG iterations, but comes with a trade-off of burdening the RMP with more, possibly unnecessary, columns.  Ultimately, we choose to return the twenty lowest reduced cost routes found.

\end{document}